\documentclass[a4paper,twoside]{article}

\usepackage{epsfig}
\usepackage{calc}
\usepackage{amssymb}
\usepackage{amstext}
\usepackage{amsmath}
\usepackage{amsthm}
\usepackage{amssymb}
\usepackage{acronym}
\usepackage{graphicx}
\usepackage{color}
\usepackage{url}
\usepackage{multicol}
\usepackage{multirow}
\usepackage{bigstrut}
\usepackage{rotating}
\usepackage{pslatex}
\usepackage[colorlinks=true,breaklinks=true,bookmarks=false]{hyperref}
\usepackage{SCITEPRESS}     

\newlength{\frameSize}

\newcommand{\vertboxs}[1]{\rotatebox{90}{\parbox{14mm}{\centering #1}}}
\newcommand{\vertboxt}[1]{\rotatebox{90}{\parbox{10mm}{\centering #1}}}

\newacro{CNN}{Convolutional Neural Network}
\newacro{CRF}{Conditional Random Field}
\newacro{FCN}{Fully Convolutional Neural Network}
\newacro{FROC}{Free-Response Receiver Operating Characteristic}
\newacro{GMM}{Gaussian Mixture Model}
\newacro{HOG}{Histogram of Oriented Gradients}
\newacro{LitW}{Logos in the Wild}
\newacro{PCA}{Principle Component Analysis}
\newacro{RPN}{Region Proposal Network}
\newacro{SIFT}{Scale Invariant Feature Transform}
\newacro{SVM}{Support Vector Machine}

\newcommand{\xie}{i.e.}
\newcommand{\xeg}{e.g.}

\newcommand{\map}{$\mathit{map}$}

\begin{document}

\title{Open Set Logo Detection and Retrieval}

\author{\authorname{Andras T\"uzk\"o\sup{1}, Christian Herrmann\sup{1,2}, Daniel Manger\sup{1}, J\"urgen Beyerer\sup{1,2}}
\affiliation{\sup{1} Fraunhofer IOSB, Karlsruhe, Germany}
\affiliation{\sup{2} Karlsruhe Institute of Technology KIT, Vision and Fusion Lab, Karlsruhe, Germany}
\email{$\lbrace$andras.tuezkoe$|$christian.herrmann$|$daniel.manger$|$juergen.beyerer$\rbrace$@iosb.fraunhofer.de}}


\abstract{
Current logo retrieval research focuses on closed set scenarios. We argue that the logo domain is too large for this strategy and requires an open set approach. To foster research in this direction, a large-scale logo dataset, called Logos in the Wild, is collected and released to the public.
A typical open set logo retrieval application is, for example, assessing the effectiveness of advertisement in sports event broadcasts. Given a query sample in shape of a logo image, the task is to find all further occurrences of this logo in a set of images or videos. Currently, common logo retrieval approaches are unsuitable for this task because of their closed world assumption. Thus, an open set logo retrieval method is proposed in this work which allows searching for previously unseen logos by a single query sample. A two stage concept with separate logo detection and comparison is proposed where both modules are based on task specific \acp{CNN}. If trained with the Logos in the Wild data, significant performance improvements are observed, especially compared with state-of-the-art closed set approaches.
}

\onecolumn \maketitle \normalsize \vfill

\section{\uppercase{Introduction}}
\label{sec:introduction}
\noindent Automated search for logos is a desirable task in visual image analysis.
A key application is the effectiveness measurement of advertisements. Being able to find all logos in images that match a query, for example, a logo of a specific company, allows to assess the visual frequency and prominence of logos in TV broadcasts. Typically, these broadcasts are sports events where sponsorship and advertisement is very common. 
This requires a flexible system where the query can be easily defined and switched according to the current task. Especially, also previously unseen logos should be found even if only one query sample is available.
This requirement excludes basically all current logo retrieval approaches because they make a closed world assumption in which all searched logos are known beforehand. Instead, this paper focuses on open set logo retrieval where only one sample image of a logo is available. 

\begin{figure}[t]
  \centering
  \includegraphics[width=\linewidth, trim=0cm 25.4cm 0cm 0cm, clip]{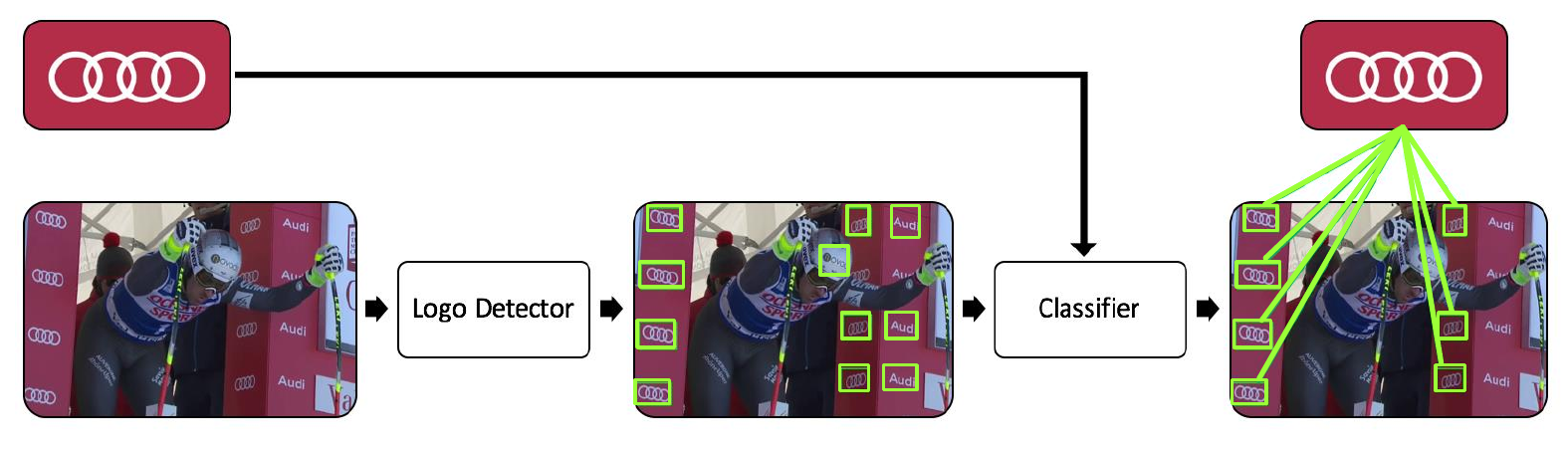}
  \caption{Proposed logo retrieval strategy.}
  \label{fig:pipeline}
\end{figure}

Consequently, a novel processing strategy for logo retrieval based on a logo detector and a feature extractor is proposed as illustrated in figure~\ref{fig:pipeline}. Similar strategies are known from other open set retrieval tasks, such as face or person retrieval~\cite{bauml2010,herrmann2015b}. Both, the detector and the extractor are task specific \acp{CNN}. For detection, the Faster R-CNN framework~\cite{ren2015} is employed and the extractor is derived from classification networks for the ImageNet challenge~\cite{deng2009}.

The necessity for open set logo retrieval becomes obvious when considering the diversity and amount of existing logos and brands\footnote{The term \textit{brand} is used in this work as synonym for a single logo class. Thus, a brand might also refer to a product or company name if an according logo exists.}. The METU trademark dataset~\cite{tursun2017} contains, for example, over half a million different brands. Given this number, a closed set approach where all different brands are pre-trained within the retrieval system is clearly inappropriate.
This is why our proposed feature extractor generates a discriminative logo descriptor, which generalizes to unseen logos, instead of a mere classification between previously known brands. The well-known high discriminative capabilities of \acp{CNN} allow to construct such a feature extractor.

One challenge for training a general purpose logo detector lies in appropriate training data. Many logo or trademark datasets~\cite{eakins1998,hoi2015,tursun2017} only contain the original logo graphic but no in-the-wild occurrences of these logos which are required for the target application. The need for annotated logo bounding boxes in the images limits the number of suitable available datasets. Existing logo datasets~\cite{joly2009,kalantidis2011,romberg2011,letessier2012,bianco2015,su2016,bianco2017} with available bounding boxes are often restricted to a very small number of brands and mostly high quality images. Especially, occlusions, blur and variations within a logo type are only partially covered.
To address these shortcomings, we collect the novel Logos in the Wild dataset and make it publicly available~\footnote{\url{http://s.fhg.de/logos-in-the-wild}}.

The contributions of this work are threefold:
\begin{itemize}
\item A novel open set logo detector which can detect previously unseen logos.
\item An open set logo retrieval system which needs only a single logo image as query.
\item The introduction of a novel large-scale in-the-wild logo dataset.
\end{itemize}

\section{\uppercase{Related Work}}
\noindent Current logo retrieval strategies are generally solving a closed set detection and classification problem. Eggert et.al. \cite{eggert2015} utilized \acp{CNN} to extract features from logos and determined their brand by classification with a set of \acp{SVM}. Fast R-CNN \cite{girshick2015} was used for the first time to retrieve logos from images by Iandola et al. \cite{iandola2015} and achieved superior results on the FlickrLogos-32 dataset \cite{romberg2011}. Furthermore, R-CNN, Fast R-CNN and Faster R-CNN were used in \cite{bao2016,oliveira2016,qi2017}. As closed set methods, all of them use the same brands both for training and for validation.

\subsection{Open Set Retrieval}
Retrieval scenarios in other domains are basically always considered open set, \xie,~samples from the currently searched class have never been seen before.
This is the case for general purpose image retrieval~\cite{sivic2003}, tattoo retrieval~\cite{manger2012} or for person retrieval in image or video data where face or appearance-based methods are common~\cite{bauml2010,weber2011,herrmann2015b}. The reason is that these in-the-wild scenarios offer usually a too large and impossible to capture variety of object classes. In case of persons, a class would be a person identity resulting in a cardinality of billions. Consequently, methods are designed and trained on a limited set of classes and have to generalize to previously unseen classes. We argue that this approach is also required for logo retrieval because of the vast amount of existing brands and according logos which cannot be captured in advance.
Typically, approaches targeting open set scenarios consist of an object detector and a feature extractor~\cite{zheng2016b}. The detector localizes the objects of interest and the feature extractor creates a discriminative descriptor regarding the target classes which can than be compared to query samples.

\subsection{Object Detector Frameworks}
Early detectors applied hand-crafted features, such as Haar-like features, combined with a classifier to detect objects in images \cite{viola2004}. 
Nowadays, deep learning methods surpass the traditional methods by a significant margin. In addition, they allow a certain level of object classification within the detector which is mostly used to simultaneously detect different object categories, such as  persons and cars \cite{sermanet2013}. 
The YOLO detector \cite{redmon2015} introduces an end-to-end network for object detection and classification based on bounding box regressors for object localization. This concept is similarly applied by the Single Shot MultiBox Detector (SSD) \cite{liu2016b}. 
The work on Faster Region-Based Convolutional Neural Network (R-CNN) \cite{ren2015} introduces a \ac{RPN} to detect object candidates in the feature maps and classifies the candidate regions by a fully connected network. 
Improvements of the Faster R-CNN are the Region-based Fully Convolutional Network (R-FCN) \cite{jifengdai2016b}, which reduces inference time by an end-to-end fully convolutional network, and the Mask R-CNN \cite{he2017}, adding a classification mask for instance segmentation.
\begin{figure*}%
\centering%
\includegraphics[width=\linewidth, trim=0cm 6.6cm 0cm 0cm, clip]{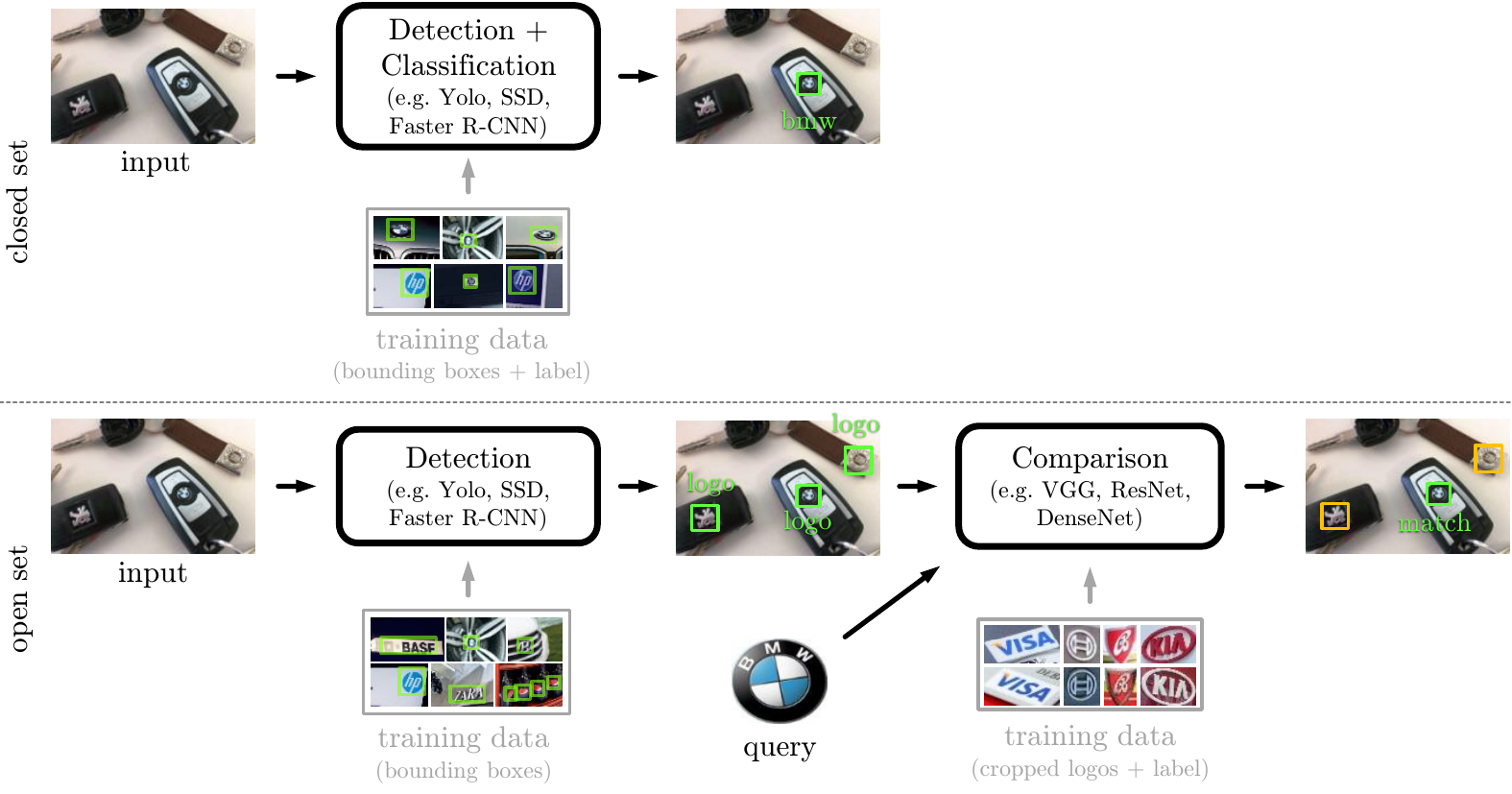}%
\caption{Comparison of closed and open set logo retrieval strategy.}%
\label{fig:openClosedSet}
\end{figure*}%

\subsection{\acs{CNN}-based Classification}
AlexNet \cite{krizhevsky2012b} was the first neural network after the conquest of \acp{SVM}, achieving impressive performance on image content classification and winning the ImageNet challenge \cite{deng2009}. It consists of five convolutional layers, each followed by a max-pooling, which counted as a very deep network at the time. 
VGG \cite{simonyan2014} follows the general architecture of AlexNet with an increased number of convolutional layers achieving better performance. 
The inception architecture~\cite{szegedy2015} proposed a multi-path network module for better multi-scale addressing, but was shortly after superseded by the Residual Networks (ResNet) \cite{he2015,he2016}. They increase network depth heavily up to 1000 layers in the most extreme configurations by additional skip connections which bypass two convolutional layers. 
The recent DenseNet \cite{huang2016} builds on a ResNet-like architecture and introduces ``dense units''. The output of these units is connected with every subsequent dense unit's input by concatenation. This results in a much denser network than a conventional feed-forward network.

\begin{table*}[t]
\centering
\caption{Publicly available in-the-wild logo datasets in comparison with the novel Logos in the Wild dataset.}
\label{tab:logoDatasets}
\begin{small}
\begin{tabular}{cl|ccc}
& \multicolumn{1}{c|}{\textbf{dataset}} & \textbf{brands} & \textbf{logo images} & \textbf{RoIs} \bigstrut[b]\\
\hline
\multirow{8}{*}{\vertboxs{public}} & BelgaLogos \cite{joly2009,letessier2012} & 37 & 1,321 & 2,697 \bigstrut[t] \\
& FlickrBelgaLogos \cite{letessier2012} & 37 & 2,697 & 2,697 \\
& Flickr Logos 27 \cite{kalantidis2011} & 27 & 810 & 1,261 \\
& FlickrLogos-32 \cite{romberg2011} & 32 & 2,240 & 3,404 \\
& Logos-32plus \cite{bianco2015,bianco2017} & 32 & 7,830 & 12,300 \\
& TopLogo10 \cite{su2016} & 10 & 700 & 863 \bigstrut[b] \\
\cline{2-5}
& combined & 80 (union) & 15,598 & 23,222 \bigstrut \\
\hline
\begin{minipage}[c]{0.1cm}\vertboxt{new}\end{minipage} & Logos in the Wild & 871 & 11,054 & 32,850 \bigstrut[t]
\end{tabular}
\end{small}
\end{table*}
\section{\uppercase{Logo Detection}}
\label{sec:detection}
\noindent The current state-of-the-art approaches for scene retrieval create a global feature of the input image. This is achieved by either inferring from the complete image or by searching for key regions and then extracting features from the located regions, which are finally fused into a global feature \cite{torii2015,arandjelovic2016,kalantidis2016}. For logo retrieval, extraction of a global feature is counterproductive because it lacks discriminative power to retrieve small objects. Additionally, global features usually include no information about the size and location of the objects which is also an important factor for logo retrieval applications.

Therefore, we choose a two-stage approach consisting of logo detection and logo classification as figure~\ref{fig:openClosedSet} illustrates for the open set case. First, the logos have to be detected in the input image. 
Since currently almost only Faster R-CNNs~\cite{ren2015} are used in the context of logo retrieval, we follow this choice for better comparability and because it offers a straightforward baseline method.
Other state-of-the-art detector options, such as SSD \cite{liu2016b} or YOLO \cite{redmon2016}, potentially offer a faster detection at the cost of detection performance \cite{huang2016b}.

Detection networks trained for the currently common closed set assumption are unsuitable to detect logos in an open set manner. By considering the output brand probability distribution, no derivation about occurrences of other brands are possible. Therefore, the task raises the need for a generic logo detector, which is able to detect all logo brands in general.

\subsubsection*{Baseline}
Faster R-CNN consists of two stages, the first being an \ac{RPN} to detect object candidates in the feature maps and the second being classifiers for the candidate regions. While the second stage sharply classifies the trained brands, the \ac{RPN} will generate candidates that vaguely resemble any of the brands which is the case for many other logos. Thus, it provides an indicator whether a region of the image is a logo or not. The trained \ac{RPN} and the underlying feature extractor network are isolated and employed as a baseline open set logo detector. 

\subsubsection*{Brand Agnostic}
The \ac{RPN} strategy is by no means optimal because it obviously has a bias towards the pre-trained brands and also generates a certain amount of false positives.
Therefore, another option to detect logos is suggested which we call the brand agnostic Faster R-CNN. It is trained with only two classes: background and logo. We argue that this solution which merges all brands into a single class yields better performance than the \ac{RPN} detector because of two reasons. 
First, in the second stage, fully connected layers preceding the output layer serve as strong classifiers which are able to eliminate false positives.
Second, these layers also serve as stronger bounding box regressors improving the localization precision of the logos.

\section{\uppercase{Logo Comparison}}
\noindent After logos are detected, the correspondences to the query sample have to be searched. 
For logo retrieval, features are extracted from the detected logos for comparison with the query sample. Then, the logo feature vector for the query image and the ones for the database are collected and normalized. Pair-wise comparison is then performed by cosine similarity.

In order to retrieve as many logos from the images as possible, the detector has to operate at a high recall. However, for difficult tasks, such as open set logo detection, high recall values induce a certain amount of false positive detections. The feature extraction step thus has to be robust and tolerant to these false positives. 

Donahue et al. suggested that \acp{CNN} can produce excellent descriptors of an input image  even in the absence of fine-tuning to the specific domain of the image \cite{donahue2015}. 
This motivates to apply a network pre-trained on a very large dataset as feature extractor. Namely, several state-of-the-art \acp{CNN} trained on the ImageNet dataset \cite{deng2009} are explored for this task. To adjust the network to the logo domain and the false positive removal, the networks are fine-tuned on logo detections. The final network layer is extracted as logo feature in all cases.

Altogether, the proposed logo retrieval system consists of a class agnostic logo detector and a feature extractor network. This setup is advantageous for the quality of the extracted logo features because the extractor network has only to focus on a specific region.
This is an improvement compared to including both logo detection and comparison in the regular Faster R-CNN framework which lacks generalization to unseen classes. We argue that the specialization in the regular Faster R-CNN to the limited number of specific brands in the training set does not cover the complexity and breadth of the logo domain. This is why a separate and more elaborate feature extractor is proposed. 

\setlength{\frameSize}{3.7cm}
\begin{figure*}%
\centering%
\includegraphics[height=\frameSize]{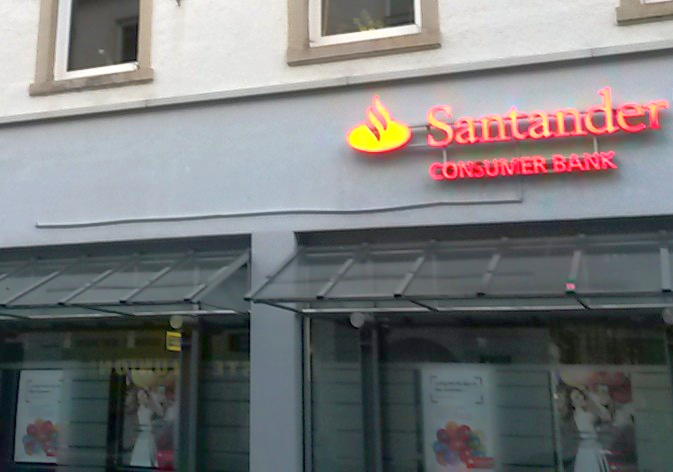}
\hfill
\includegraphics[height=\frameSize]{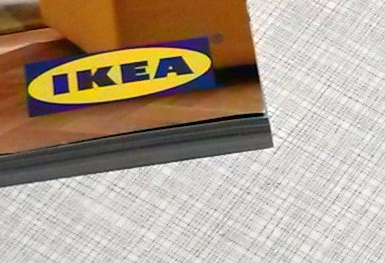}
\hfill
\includegraphics[height=\frameSize]{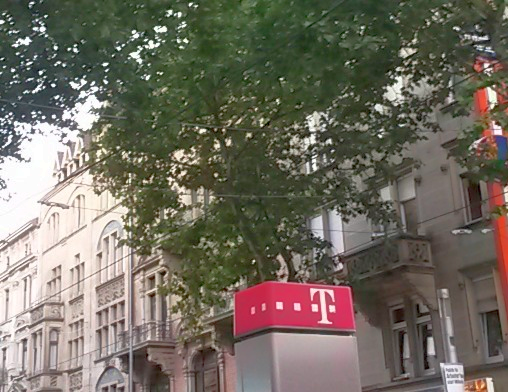}%
\\
\vspace{1.5mm}%
\includegraphics[height=\frameSize]{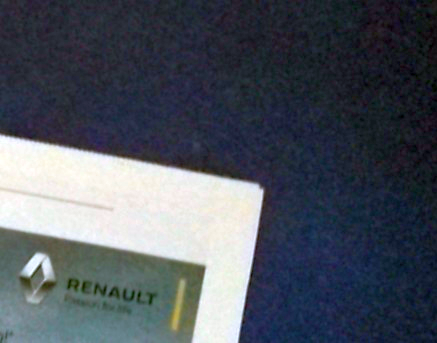}
\hfill
\includegraphics[height=\frameSize]{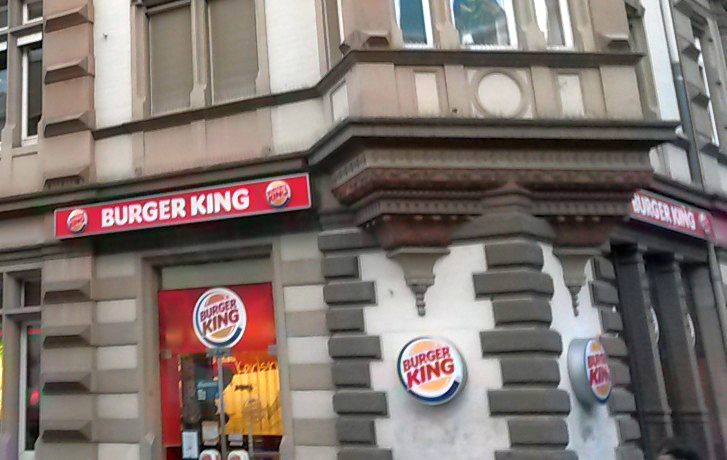}
\hfill
\includegraphics[height=\frameSize]{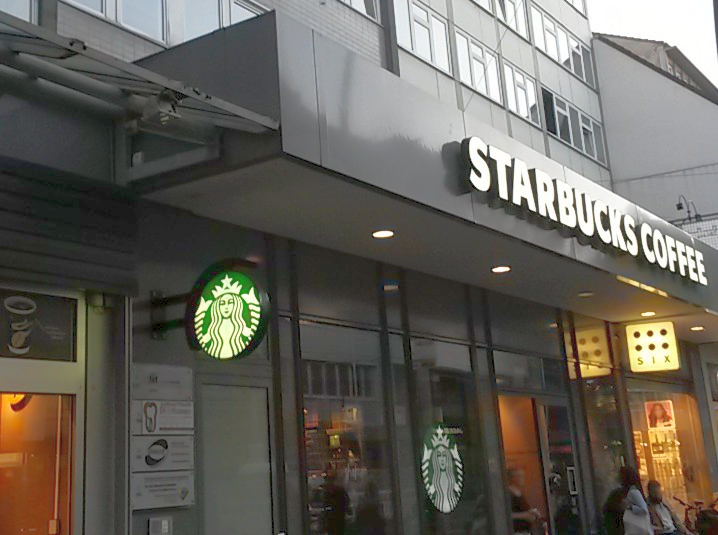}
\caption{Examples from the collected Logos in the Wild dataset.}%
\label{fig:logoSamples}
\end{figure*}%
\section{\uppercase{Logo Dataset}}
\noindent To train the proposed logo detector and feature extractor, a novel logo dataset is collected to supplement publicly available logo datasets. A comparison to other public in-the-wild datasets with annotated bounding boxes is given in table~\ref{tab:logoDatasets}.
The goal is an in-the-wild logo dataset with images including logos instead of the raw original logo graphics. In addition, images where the logo represents only a minor part of the image are preferred. See figure~\ref{fig:logoSamples} for a few examples of the collected data.
Following the general suggestions from \cite{bansal2017}, we target for a dataset containing significantly more brands instead of collecting additional image samples for the already covered brands. This is the exact opposite strategy than performed by the Logos-32plus dataset.
Starting with a list of well-known brands and companies, an image web search is performed. Because most other web collected logo datasets mainly rely on Flickr, we opt for Google image search to broaden the domain. Brand or company names are searched directly or in combination with a predefined set of search terms, \xeg, `advertisement', `building', `poster' or `store'. 
\begin{figure}%
\centering%
\includegraphics[width=\linewidth, trim=0cm 9.9cm 0cm 0cm, clip]{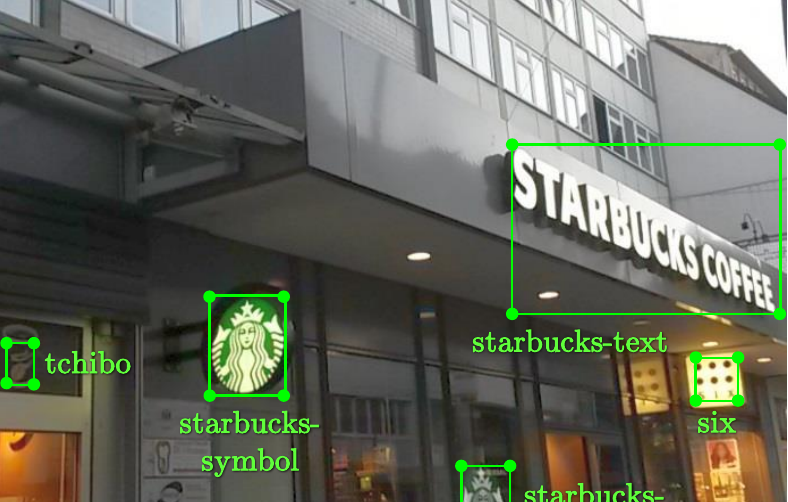}%
\caption{Annotations differentiate between textual and graphical logos.}%
\label{fig:annotatedSample}
\end{figure}%

\begin{figure}%
\centering%
\includegraphics[width=\linewidth, trim=0cm 11cm 0cm 0cm, clip]{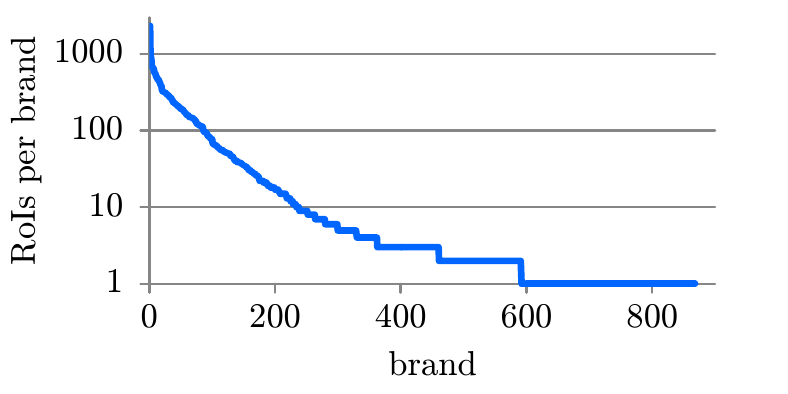}%
\caption{Distribution of number of RoIs per brand.}%
\label{fig:brandDistribution}
\end{figure}

\begin{figure}%
\centering%
\includegraphics[width=\linewidth, trim=0cm 11cm 0cm 0cm, clip]{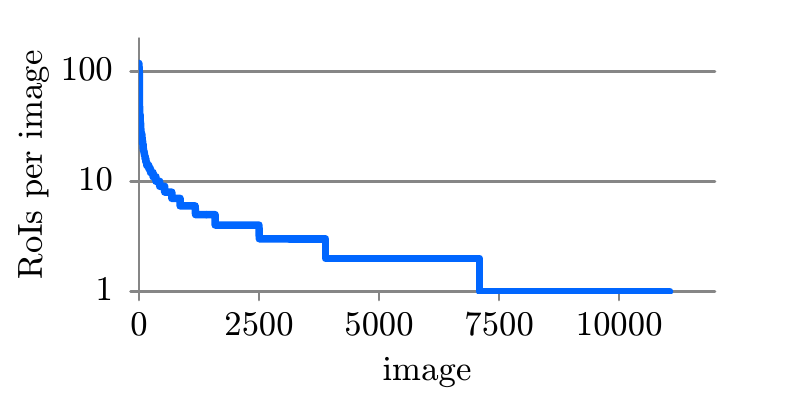}%
\caption{Distribution of number of RoIs per image.}%
\label{fig:logoDistribution}
\end{figure}

For each search result, the first $N$ images are downloaded, where $N$ is determined by a quick manual inspection to avoid collecting too many irrelevant images. 
After removing duplicates, this results in 4~to 608~images per searched brand. These images are then one-by-one manually annotated with logo bounding boxes or sorted out if unsuitable.
Images are considered unsuitable if they contain no logos or fail the in-the-wild requirement, which is the case for the original raw logo graphics. Taken pictures of such logos and advertisement posters on the other hand are desired to be in the dataset. 
Annotations distinguish between textual and graphical logos as well as different logos from one company as exemplary indicated in figure~\ref{fig:annotatedSample}.
Altogether, the current version of the dataset, contains 871 brands with 32,850 annotated bounding boxes. 238 brands occur at least 10 times. An image may contain several logos with the maximum being 118 logos in one image. The full distributions are shown in figures~\ref{fig:brandDistribution} and~\ref{fig:logoDistribution}. 

The collected Logos in the Wild dataset exceeds the size of all related logo datasets as shown in table~\ref{tab:logoDatasets}. Even the union of all related logo datasets contains significantly less brands and RoIs which makes Logos in the Wild a valuable large-scale dataset.
As the annotation is still an ongoing process, different dataset revisions will be tagged by version numbers for future reference. Note that the numbers in table~\ref{tab:logoDatasets} are the current state (v2.0) whereas detector and feature extractor training used a slightly earlier version with numbers given in table~\ref{tab:trainTestStatistics} (v1.0) because of the required time for training and evaluation.

\section{\uppercase{Experiments}}
\noindent The proposed method is evaluated on the test set benchmark of the public FlickrLogos-32 dataset including the distractors. Additional application specific experiments are performed on an internal dataset of sports event TV broadcasts. 
The training set consists of two parts. The union of all public logo datasets as listed in table~\ref{tab:logoDatasets} and the novel \ac{LitW} dataset. For a proper separation of train and test data, all brands present in the FlickrLogos-32 test set are removed from the public and \ac{LitW} data. Ten percent of the remaining images are set aside for network validation in each case. This results in the final training and test set sizes listed in table~\ref{tab:trainTestStatistics}.
\begin{table}[t]
\centering
\begingroup	
\setlength{\tabcolsep}{6pt}
\caption{Train and test set statistics.}
\label{tab:trainTestStatistics}
\begin{small}
\begin{tabular}{cl|cc}
\textbf{phase} & \multicolumn{1}{c|}{\textbf{data}} & \multicolumn{1}{c}{\textbf{brands}} & \multicolumn{1}{c}{\textbf{RoIs}} \bigstrut[b]\\
\hline
\multirow{2}[2]{*}{train} & public & 47    & 3,113 \bigstrut[t]\\
      & public+\ac{LitW} v1.0 & 632   & 18,960 \bigstrut[b]\\
\hline
test  & FlickrLogos-32 test & 32    & 1,602 \bigstrut[t]\\
\end{tabular}%
\end{small}
\endgroup
\end{table}

In the first step, the detector stage alone is assessed. Then, the combination of detection and comparison for logo retrieval is evaluated. 
Detection and matching performance is measured by the \ac{FROC} curve~\cite{miller1969} which denotes the detection or detection and identification rate versus the number of false detections.
In all cases, the \acp{CNN} are trained until convergence. Due to the diversity of applied networks and differing dataset sizes, training settings are numerous and optimized in each case with the validation data. Convergence occurs after 200 to 8,000 training iterations with a varying batch-size of 1 for the Faster R-CNN detector, 7 for the DenseNet161, 18 for the ResNet101 and 32 for the VGG16 training due to GPU memory limitation.

\subsection{Detection}
\begin{figure}%
\centering%
\includegraphics[width=\linewidth, trim=0cm 9cm 0cm 0cm, clip]{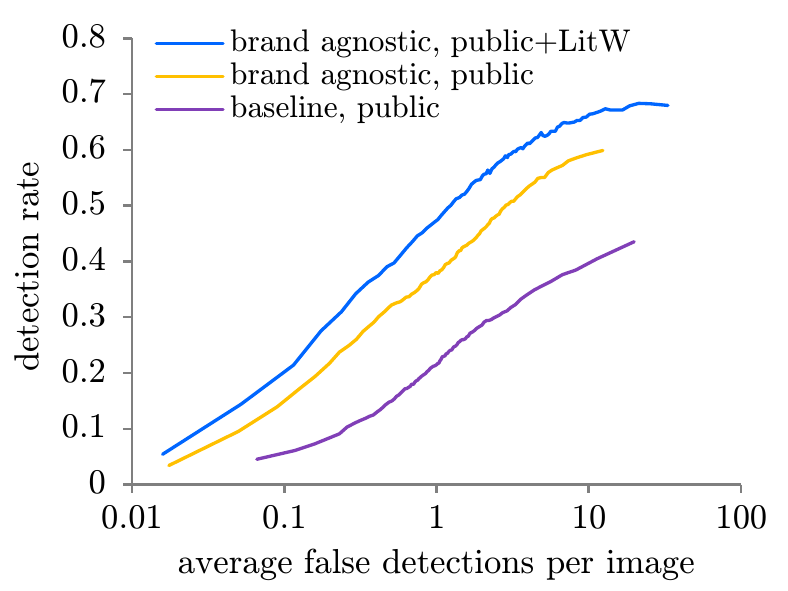}%
\caption{Detection FROC curves for the FlickrLogos-32 test set.}%
\label{fig:detectionFroc}
\end{figure}

As indicated in section~\ref{sec:detection}, the baseline is the state-of-the-art closed set logo retrieval method from~\cite{su2016} which is trained on the public data and naively adapted to open set detection by using the \ac{RPN} scores as detections
The proposed brand agnostic logo detector is first trained on the same public data for comparison. All Faster R-CNN detectors are based on the VGG16 network.
The results in figure~\ref{fig:detectionFroc} indicate that the proposed brand agnostic strategy is superior by a significant margin. 

Further improvement is achieved by combining the public training data with the novel logo data.
Adding \ac{LitW} as additional training data improves the detection results with its large variety of additional training brands. 
This confirms findings from other domains, such as face analysis, where wider training datasets are preferred over deeper ones~\cite{bansal2017}. This means it is better to train on additional different brands than on additional samples per brand.
As direction for future dataset collection, this suggests to focus on additional brands.

\subsection{Retrieval}
For the retrieval experiments, the Faster R-CNN based state-of-the-art closed set logo retrieval method from the previous section serves again as baseline. Now the full network is applied and the logo class probabilities of the second stage are interpreted as feature vector which is then used to match previously unseen logos.
For the proposed open set strategy, the best logo detection network from the previous section is used in all cases. Detected logos are described by the feature extraction network outputs where three different state-of-the-art classification architectures, namely VGG16~\cite{simonyan2014}, ResNet101~\cite{he2015} and DenseNet161~\cite{huang2016}, serve as base networks. All networks are pretrained on ImageNet and afterwards fine-tuned either on the public logo train set or the combination of the public and the \ac{LitW} train data.

\begin{figure}%
\centering%
\includegraphics[width=\linewidth, trim=0cm 7.4cm 0cm 0cm, clip]{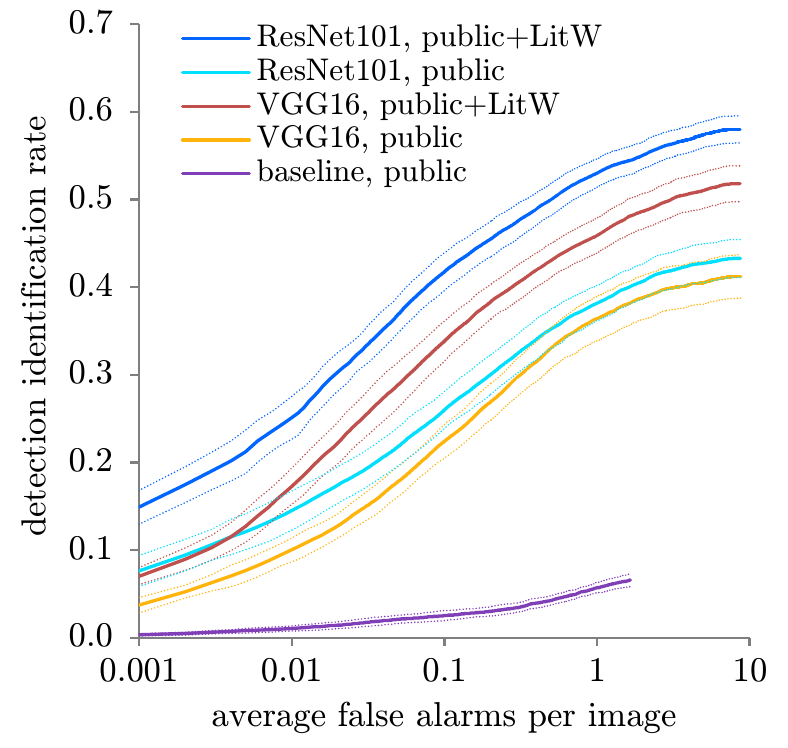}%
\caption{Detection+Classification FROC curves for the FlickrLogos-32 test set. Including dashed indicators for one standard deviation. DenseNet results are omitted for clarity, refer to table~\ref{tab:mapFlickr} for full results.}%
\label{fig:classificationFroc}
\end{figure}%
\subsubsection*{FlickrLogos-32}
In ten iterations, each of the ten FlickrLogos-32 train samples for each brand serves as query sample. This allows to assess the statistical significance of results similar to a 10-fold-cross-validation strategy. Figure~\ref{fig:classificationFroc} shows the FROC results for the trained networks including indicators for the standard deviation of the measurements. The detection identification rate denotes the amount of ground truth logos which are correctly detected and are assigned the correct brand.
While the baseline method is only able to find a minor amount of the logos, our best performing approach is able to correctly retrieve 25 percent of the logos if tolerating only one false alarm every 100 images.
As expected, the more recent network architectures provide better results. Also, including the \ac{LitW} data in the training yields a significant boost in performance. Specifically, the larger training dataset has a larger impact on the performance than a better network architecture. 

Table~\ref{tab:mapFlickr} compares our open set results with closed set results from the literature in terms of the mean average precision (\map). 
\begin{table}[t]
\centering
\begingroup	
\setlength{\tabcolsep}{6pt}
\caption{FlickrLogos-32 test set retrieval results.}
\label{tab:mapFlickr}
\begin{small}
\begin{tabular}{crc}
\textbf{setting} & \multicolumn{1}{c}{\textbf{method}} & \textbf{\map} \bigstrut[b]\\
\hline
\multirow{7}[2]{*}{\begin{sideways}open set\end{sideways}} & baseline, public~\cite{su2016} & 0.036 \bigstrut[t]\\
      & \textit{VGG16, public} & 0.286 \\
      & \textit{ResNet101, public} & 0.327 \\
      & \textit{DenseNet161, public} & 0.368 \\
\cline{2-3}      
      & \textit{VGG16, public+\ac{LitW}} & 0.382 \bigstrut[t]\\
      & \textit{ResNet101, public+\ac{LitW}} & \textbf{0.464} \\
      & \textit{DenseNet161, public+\ac{LitW}} & 0.448  \bigstrut[b]\\      
\hline
\multirow{4}[1]{*}{\begin{sideways}closed set\end{sideways}} & BD-FRCN-M \cite{oliveira2016} & 0.735 \bigstrut[t]\\	
      & DeepLogo \cite{iandola2015} & 0.744 \\ 
      & Faster-RCNN \cite{su2016} & 0.811 \\
      & Fast-M \cite{bao2016} & \textbf{0.842} \\
\end{tabular}%
\end{small}
\endgroup
\end{table}
We achieve more than half of the closed set performance in terms of \map~with only one sample for a brand at test time instead of dozens or hundreds of brand samples at training time. 
Having only a single sample is a significant harder retrieval task on FlickrLogos-32 than closed set retrieval because logo variations within a brand are uncovered by this single sample. The test set includes such logo variations to a certain extent which requires excellent generalization capabilities if only one query sample is available.

In addition, our approach is not limited to the 32 FlickrLogos brands but generalizes with a similar performance to further brands. In contrast, the closed set approaches hardly generalize as is shown by the baseline open set method which is based on the second best closed set approach. The only difference is the training on out-of-test brands for the open set task.

\subsubsection*{SportsLogos}
\begin{figure*}%
\centering%
\includegraphics[width=\linewidth, trim=0cm 0cm 0cm 0.5cm, clip]{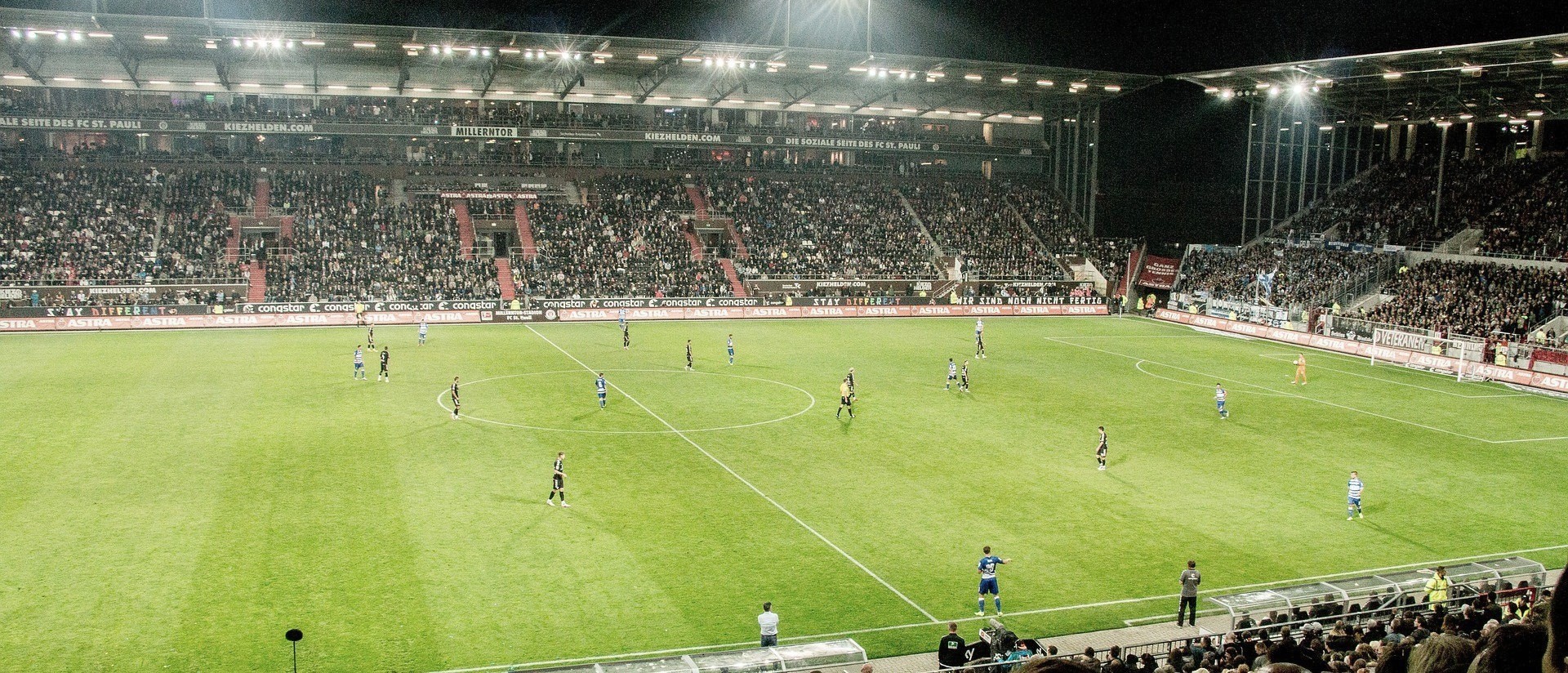}%
\caption{Example football scene with small logos in the perimeter advertising.}%
\label{fig:footballSample}
\end{figure*}

\begin{figure}%
\centering%
\includegraphics[width=\linewidth, trim=0cm 8.5cm 0cm 0cm, clip]{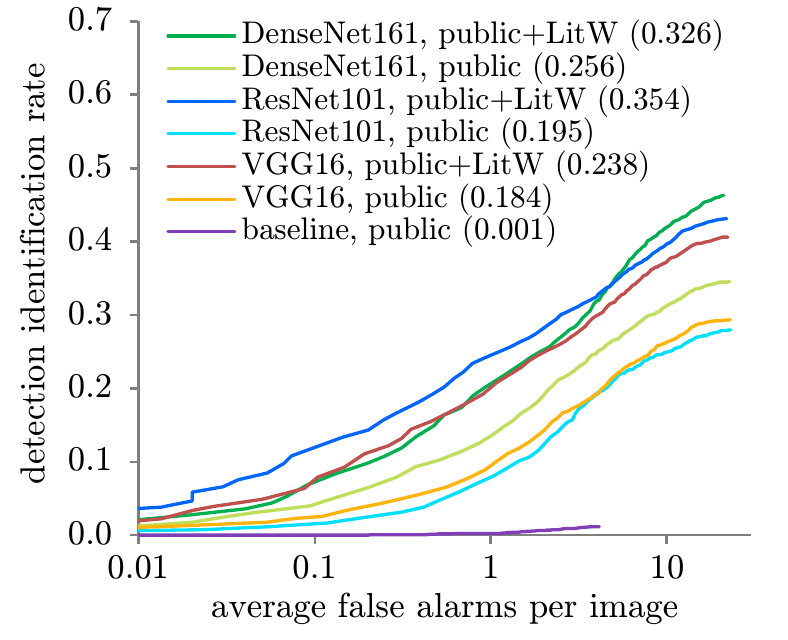}%
\caption{Detection+Classification FROC curves for the Sports\-Logos test set, \map~is given in brackets.}%
\label{fig:classificationFrocFootball}
\end{figure}

In addition to public data, target domain specific experiments are performed on TV broadcasts of sports events. In total, this non-public test set includes 298 annotated frames with 2,348 logos of 40 brands. In comparison to public logo datasets, the logos are usually significantly smaller and cover only a tiny fraction of the image area as illustrated in figure~\ref{fig:footballSample}. Besides perimeter advertising, logos on clothing or equipment of the athletes and TV station or program overlays are the most occurring logo types.
Overall, the results in this application scenario are slightly worse than in the FlickrLogos-32 benchmark with a drop in \map~from 0.464 to 0.354 for the best performing method, as indicated in figure~\ref{fig:classificationFrocFootball}. The baseline approach takes the largest performance hit showing that closed set approaches not only generalize badly to unseen logos but also to novel domains. In contrast, the proposed open set strategy shows a relatively stable cross-domain performance. Training with \ac{LitW} data again improves the results significantly.

\section{\uppercase{Conclusions}}
\label{sec:conclusion}
\noindent The limits of closed set logo retrieval approaches motivate the proposed open set approach. By this, generalization to unseen logos and novel domains is improved significantly in comparison to a naive extension of closed set approaches to open set configurations. 
Due to the large logo variety, open set logo retrieval is still a challenging task where trained methods benefit significantly from larger datasets. The lack of sufficient data is addressed by introduction of the large-scale Logos in the Wild dataset. Despite being bigger than all other in-the-wild logo datasets combined, dataset sizes should probably be scaled even further in the future. Adding the Logos in the Wild data in the training improves the mean average precision from 0.368 to 0.464 for open set logo retrieval on FlickrLogos-32.


\bibliographystyle{ieee}
{\small
\bibliography{Biblio}}

\vfill
\end{document}